\documentclass[11pt,a4paper]{article}
\usepackage[hyperref]{acl2021}
\usepackage{times}
\usepackage{latexsym}

\usepackage{microtype}

\aclfinalcopy 
\def\aclpaperid{2879} 

\usepackage{todonotes}

\usepackage{enumitem}
\usepackage{booktabs}
\usepackage{multirow}
\usepackage{amsmath}
\usepackage{amsfonts}
\usepackage{fontawesome}
\usepackage[ruled,vlined]{algorithm2e}
\usepackage{arydshln}
\usepackage[procnames]{listings}
\usepackage{multicol}
\usepackage[title]{appendix}
\usepackage{makecell}
\usepackage{color, colortbl}
\definecolor{Gray}{gray}{0.9}

\usepackage{longtable}
\usepackage{todonotes}

\usepackage{adjustbox}

\title{Biomedical Interpretable Entity Representations}

\author{
Diego Garcia-Olano$^{1,2}$ 
Yasumasa Onoe$^2$ 
Ioana Baldini$^1$ \\
\textbf{Joydeep Ghosh}$^2$ 
\textbf{Byron C. Wallace}$^3$ 
\textbf{Kush R. Varshney}$^1$\\
$^1$IBM Research,
$^2$University of Texas at Austin,
$^3$Northeastern University\\
diegoolano@gmail.com,
yasumasa@utexas.edu,
ioana@us.ibm.com,\\
ghosh@ece.utexas.edu,
b.wallace@northeastern.edu,
krvarshn@us.ibm.com
}

\date{}

\begin{document}
\maketitle
\begin{abstract}
Pre-trained language models induce dense entity representations that offer strong performance on entity-centric NLP tasks, but such representations are not immediately interpretable.
This can be a barrier to model uptake in important domains such as biomedicine. There has been recent work on general interpretable representation learning \cite{onoe-durrett-2020-interpretable}, but these domain-agnostic representations do not readily transfer to the important domain of biomedicine. 
In this paper, we create a new entity type system and training set from a large corpus of biomedical texts by mapping entities to concepts in a medical ontology, and from these to Wikipedia pages whose categories are our types. 
From this mapping we derive \emph{Biomedical Interpretable Entity Representations} (BIERs), in which dimensions correspond to fine-grained entity types, and values are predicted probabilities that a given entity is of the corresponding type. 
We propose a novel method that exploits BIER's final sparse and intermediate dense representations to facilitate model and entity type debugging. We show that BIERs achieve strong performance in biomedical tasks including named entity disambiguation and entity label classification, and we provide error analysis to highlight the utility of their interpretability, particularly in low-supervision settings.  
Finally, we provide our induced 68K biomedical type system, the corresponding 37 million triples of derived data used to train BIER models and our best performing model.

\end{abstract}

\section{Introduction}

In modern NLP systems, entities are embedded in the same dense vector space as words using vectors from pre-trained (masked) language models ~\cite{Jacob_Devlin_19} that yield contextualized embeddings of tokens. 
These representations are used as inputs for downstream models built for particular tasks.
One issue with such learned representations is that we do not actually know what information they encode. 
Recent work has shown that deep pre-trained models implicitly learn factual knowledge about entities \cite{Fabio_Petroni_19, Adam_Roberts_20}, but the embeddings that they provide do not explicitly maintain representations of this knowledge (i.e., the dimensions in learned representations have no \emph{a priori} semantics); consequently, are not directly interpretable.
This has motivated the design of knowledge \emph{probing tasks} to measure a factual knowledge implicit in embeddings \cite{Fabio_Petroni_19, Nina_Poerner_19}. 

Recent work \cite{onoe-durrett-2020-interpretable} has proposed learning interpretable entity representations using an entity typing model and corresponding fine-grained type system that accepts an entity mention and its context. The output represents a high-dimensional sparse embedding whose values correspond to the model's (independently) predicted probabilities that the entity possesses the respective properties defined by the fine-grained type system.

This past work proposed general domain pre-trained Transformer-based~\cite{Ashish_Vaswanir_17} entity typing models trained on Wikipedia or the ultra-fine entity typing system~\cite{Eunsol_Choi_18}, yielding 60k and 10k dimensional embeddings, respectively, which can then be used directly in downstream tasks. 
Such representations can achieve strong results 
without learning task specific representations.
Thus, in addition to providing interpretability, such representations may be particularly useful for tasks with limited supervision.

Such interpretable entity representations for text can be valuable in domains such as biomedicine, because they afford model transparency which may help with model debugging, or simply to instill confidence in model outputs. 
For example, if one defines a linear layer on top of entity-type representations, learned coefficients are interpretable as weights assigned to specific entity types. 
One could debug an incorrect prediction by inspecting the induced representation for potentially erroneous types assigned to it.
This sort of insight is particularly important in biomedical NLP, given the potential sensitivity of the tasks in the domain, and the high-level expertise of the `end-users'.

Motivated by these observations, we extend \cite{onoe-durrett-2020-interpretable} to learn sparse Biomedical Interpretable Entity Representations (BIERs) in which values encode predicted probabilities of an entity belonging to a type from a fine-grained entity type system.  
Starting from a corpus of PubMed\footnote{A repository of biomedical literature: \url{http://www.pubmed.gov/}.} articles on cancer and drugs as our training data, we induce an entity type system by mapping entities in the articles to their associated UMLS concepts, and then mapping the concepts to Wikipedia pages whose categories we use as our types. 

We show that learning a typing model on top of such a system realizes strong performance on a variety of biomedical tasks including named entity disambiguation (NED) and entity label classification using simple cosine similarity or Euclidean distance based methods, and we provide an analysis of the results from an interpretabilty perspective. 
In addition, we propose a simple technique that facilitates debugging and provides a mechanism by which to improve model performance by exploiting both the proposed sparse interpretable type representations and their internal underlying dense counterparts.  
Finally, we introduce and release a new medical-centric Wikipedia dataset based on~\cite{rosenthal-etal-2019-leveraging} for use in the task of biomedical NED.

Our specific contributions\footnote{Code and datasets available at \url{http://github.com/diegoolano/biomedical_interpretable_entity_representations}} are as follows: 
\begin{itemize}
    \item We create (and will release) a biomedical entity typing system comprising Wikipedia Categories from pages mapped to UMLS concepts linked to PubMed article entities and learn a model that produces sparse entity representations in which dimensions are imbued with known semantics. 
    We show that these achieve strong performance on biomedical NED and entity label classification tasks. 
    \item We conduct an interpretability analysis and demonstrate a new debugging method that uses the proposed representation's performance on downstream tasks to gain insights into the entity typing model and system.
    \item We release a medical literature centric Wikipedia dataset for use in the task of biomedical NED.
\end{itemize}

\section{Background: Interpretable Entity Model}
We first review the interpretable entity model architecture we extend from \cite{onoe-durrett-2020-interpretable}. 


\begin{figure}[!t]
    \centering
    \includegraphics[width=1.0\linewidth]{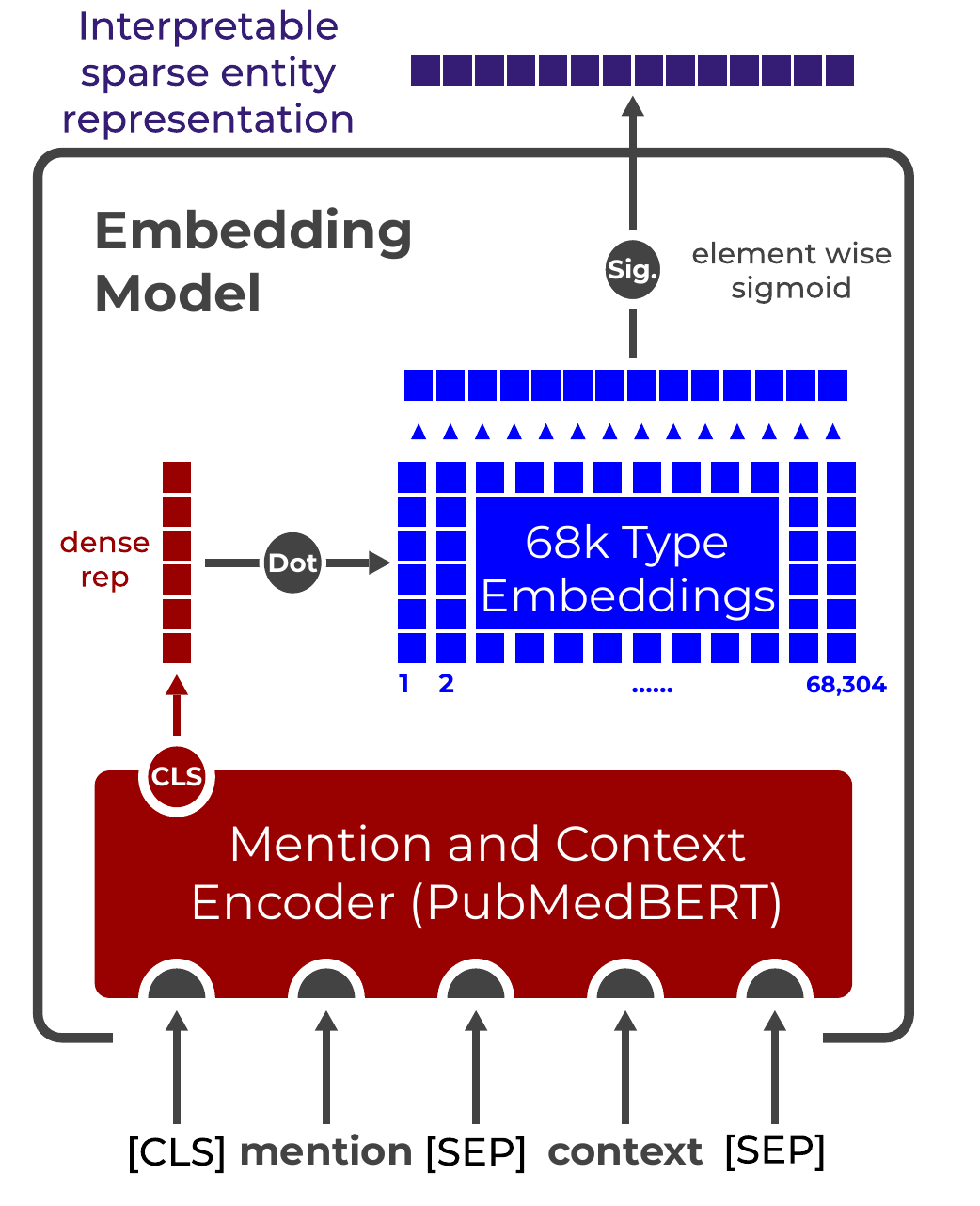}
    \caption{Model architecture from \cite{Yasumasa_Onoe_19} using our 68k biomedical entity type system. A BERT based encoder embeds a mention and context and the output entity embedding  contains probabilities for each type. }
    \label{fig:model}
\end{figure}


Let $s = (w_1, ..., w_N)$ denote a sequence of input context words, $m = (w_i, ..., w_j)$ denote an entity mention span in $s$, and $\mathbf{t} \in  [0,1]^{|\mathcal{T}|}$ denote a vector whose values are predicted probabilities corresponding to fine-grained entity types $\mathcal{T}$ from a predefined type system with higher values identifying types most pertaining to $m$ and $s$ .

Given a labeled dataset $\mathcal{D} = \{(m, s, \mathbf{t}^*)^{(1)}, ... , (m, s, \mathbf{t}^*)^{(k)} \}$ the objective is to learn parameters $\theta$ of a function $f_\theta$ that maps the mention $m$ and its context $s$ to a vector $\mathbf{t}$ that captures salient features of the entity mention within its context. 
The basic idea is that the resultant entity embeddings $\mathbf{t}$ (wherein individual dimensions have explicit semantics) can be used as embeddings in downstream tasks, for example by using basic similarity measures such as dot products or cosine similarities.\footnote{Fine-tuning the representations would destroy their interpretability.}

The simple model $f_\theta$ that produces these embeddings is shown in Figure~\ref{fig:model}. 
First, a BERT-based encoder~\citep{ Jacob_Devlin_19} maps inputs $m$ and $s$ to an intermediate dense vector representation. Specifically, the encoder takes as input a token sequence formatted as $\mathbf{x} =$ {\tt[CLS]} $m$ {\tt[SEP]} $s$ {\tt[SEP]}, where the mention $m$ and context $s$ are segmented into WordPiece tokens \citep{Yonghui_Wu_16}. 
The hidden vector output corresponding to the {\tt [CLS]} token can be treated as the intermediate dense mention and context representation: $\mathbf{h}_{\texttt{[CLS]}} = \textsc{BertEncoder}(\mathbf{x})$.  

A type embedding layer then projects this intermediate representation to a vector whose dimensions correspond to the entity types $\mathcal{T}$ using a single linear layer whose parameters may be viewed as a matrix of type embeddings $\mathbf{E} \in  \mathbb{R}^{|\mathcal{T}| \times d}$, where $d$ is the dimension of the mention and context representation $\mathbf{h}_{\texttt{[CLS]}}$.  
Finally, we apply a sigmoid function to each unnormalized score in the vector to obtain the 
predicted probabilities that form our entity representation $\mathbf{t}$ (top of Figure \ref{fig:model}). 
We obtain these output probabilities $\mathbf{t}$ by multiplying $\mathbf{E}$ by $\mathbf{h}_{\texttt{[CLS]}}$, followed by an element-wise sigmoid function: $\mathbf{t} = \sigma \left(\mathbf{E} \cdot
\mathbf{h}_{\texttt{[CLS]}}\right)$



Following~\citet{Eunsol_Choi_18}, the training loss we minimize is a sum of binary cross-entropy losses over all entity types $\mathcal{T}$ over all training examples $\mathcal{D}$. That is, we treat each type prediction for each example as an independent binary decision, with shared parameters in the BERT encoder. Our loss $\mathcal{L}$ is:
 \begin{equation*}
 \label{loss1}
 -\sum_{i}\sum_{j} t^*_{ij} \cdot \log (t_{ij}) + (1 - t^*_{ij}) \cdot \log (1 - {t_{ij}}),\\
 \end{equation*}
where $i$ are the data indices, $j$ are indices over types, $t_{ij}$ is the $j$th component of $\mathbf{t_i}$, and $t^*_{ij}$ is the $j$th component of $\mathbf{t_i}^*$ that takes the value 1 if the $j$th type applies to the current entity mention. We fine-tune all parameters in BERT and the type embedding matrix.

\section{Biomedical Interpretable Entity Representations}

\paragraph{Biomedical Entity Typing} 
To train an interpretable entity embedding model tailored specifically for biomedical tasks, we must first construct a suitable biomedical entity type system and dataset.
PubMed indexes over 30 million biomedical citations across a wide range of topics. 
To curate a topically focused set of literature, we first used the PubTator tool \cite{10.1093/nar/gkz389} to query PubMed for articles related to drugs used as treatment for cancer; this yielded 461,404 unique citations (titles and abstracts).\footnote{We selected the topic of cancer because our work is motivated by a larger project aimed at finding existing evidence that supports repurposing generic drugs for cancer.} 

We used an off the shelf NER tagger available in {\tt SciSpacy} \cite{neumann-etal-2019-scispacy} to identify entity spans within abstracts, and used the {\tt Entity Linker} component to link those entities to concept unique IDs (CUIDs) within the Unified Medical Language System (UMLS) ontology\footnote{UMLS defines around 3 million concepts from a combined 200 source ontologies.  
Concepts may be identified as having one or more of 127 semantic types which can be used to place them into groupings such as diseases or drugs.}.  


\begin{figure}[!t]
    \centering
    \includegraphics[width=1.0\linewidth]{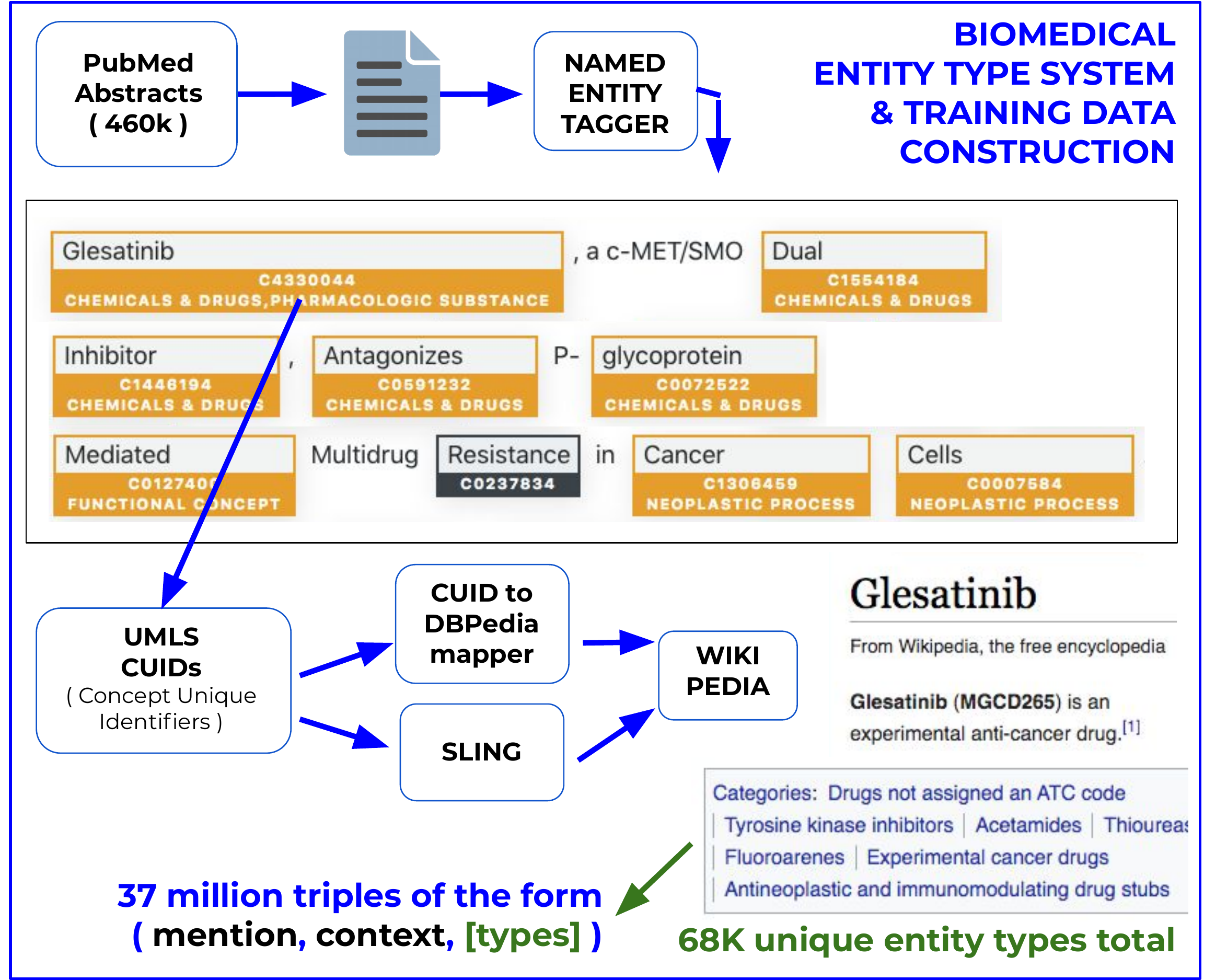}
    \caption{Biomedical Entity Type System and dataset construction. Appendix Fig \ref{fig:big_pub} contains example output.}
    \label{fig:BIERdiagram}
\end{figure}


Next we had to decide on the specific entity type system to use, i.e., the set of labels to attach to entities, and chose Wikipedia as our knowledge base.
We used this general knowledge base instead of a specialized ontology (for example, MeSH or SNOWMED CT) primarily because it yielded (many) more diverse entity types per mention, comparatively.

To connect UMLS concepts to Wikpedia pages we use the mapping from~\citet{Cuzzola2018UMLSTD}, which is accurate but incomplete: It provides exact wikipage matches for 221,690 concepts and ``close matches" for 26,276 of them, out of a possible 3 million concepts in UMLS. 
For concepts for which no exact or close match was found, we used SLING~\cite{46431}, a framework for frame semantic parsing which allows for querying and resolving wikipages given a search string (in our case, mention surface forms). 
For high confidence exact or close matches, we return the set of categories found for their combined results. While these results can be slightly noisy, they mostly lead to satisfactory performance.

We filter the entity mentions that compose our final set, as follows. 
If multiple concept CUIDs are found for a given entity, we include the highest scoring matches within two points of each other provided they all exceed a minimum score threshold of 0.8;\footnote{This is the default threshhold set in {\tt SciSpacy} for concept candidate inclusion.} 
Additionally, we only include results that are linked to at least one concept CUID and where an associated Wiki link was mapped to directly via \citet{Cuzzola2018UMLSTD} or via SLING. A schematic of this process is shown in Figure~\ref{fig:BIERdiagram}. An example working through the entity filtering process is shown in the text of Appendix A.  In the end about 12.5\% of the mappings from PubMed mentions to Wikipedia categories come via SLING.

After processing, linking and filtering the corpus of PubMed abstracts, we were able to extract 37,357,141 triples of the form (mention, context, [list of categories]). This list of triples contains 68,304 unique categories which we use as the entity type system for training BIERs. Appendix ~\ref{tab:types} contains a list of the top 100 entity types that appear over these articles and Appendix \ref{fig:bier_hist} shows a histogram of entity types per mention.  As one contribution, we will release this set of derived triples. 

To assess the quality of this dataset, we chose 500 triples at random and asked 4 experts (researchers in biomedicine and ML) to score them on a Likert scale from 1 (low) to 5 (high) for accuracy. Experts assessed how well a PubMed mention from a context sentence maps onto a Wikipedia URL. Average expert scores for the triples were [4.01, 4.13, 4.18, 4.20] (overall mean of 4.13) out of 5. The Fleiss-Kappa score which measures inter-annotator agreement was strong at .69.  Additionally 77\% of scores are $>=$ 4, and for 93\% of the examples at least 3/4 experts agree (73\% have unanimous agreement). 

\paragraph{BIER entity typing model training and test results}

We split our derived dataset of biomedical triples into train, validation, and test sets of sizes 31,340,000, 376,071, and 5,641,070, respectively. For comparison, the total data size used by \citet{onoe-durrett-2020-interpretable} is 6.1 million and based on the most popular categories of Wikipedia whereas ours only uses categories on pages linked to UMLS.  

We trained different BIER models using variants of BERT as an encoder for mentions and contexts. Specifically we considered BioBERT~\cite{10.1093/bioinformatics/btz682}, SciBERT~\cite{Beltagy-etal-2019-scibert} and BLURB~\cite{gu2020domainspecific} (we will refer to this as PubMedBERT), which constitute the current state of the art for many biomedical tasks.
We compute entity typing macro F1 using development examples to check model convergence and use the hyperparameters from \citet{onoe-durrett-2020-interpretable}. 

\paragraph{Debugging BIERs by combining dense and sparse embeddings} 
We propose a technique for debugging using BIER representations that is in part inspired by prior work that used intermediate layer representations of training examples as additional features \cite{Papernot2018DeepKN}. 
Specifically, we propose to debug BIER performance on downstream tasks by examining instances where dense and sparse representations yield different outputs. 
For each example, BIER models produce an intermediate dense  $\mathbf{h}_{\texttt{[CLS]}}$ and interpretable sparse output embeddings $t$ (red and purple, respectively, in Figure ~\ref{fig:model}). We will refer to the two seperate models which use these dense and sparse BIERs embeddings for downstream tasks as {\tt BIERDense} and {\tt BIERSparse} respectively. 

After performing inference initially, we gather all test examples where the {\tt BIERDense} makes a correct prediction but {\tt BIERSparse} does not and we place their mention values into a set $\mathcal{Z}$. 
Additionally, as a diagnostic measure, we consider an `oracle' approach in which we use the {\tt BIERDense} prediction for all instances in $\mathcal{Z}$, and the {\tt BIERSparse} output otherwise.
The intuition is that $\mathcal{Z}$ contains examples for which the intermediate dense embeddings better represent a mention-context than the more interpretable sparse output embeddings from the BIER model. 

Because the sparse embeddings are interpretable, this analysis affords fine-grained analysis of which entity types lead to incorrect predictions by the sparse model (but correct predictions using a dense representation). 
This diagnostic can be used as a benchmark for how well the model could have done had the entity typing model's output better represented the mention-context, or if the model had known to fallback to using the intermediate dense embedding; the former case might be ameliorated via more supervised examples or changes to the type system while the latter could motivate a dynamic approach to making predictions that is a function of model confidence. 
We show results and analysis using these methods in Section 5.

\section{Experimental Setup}

To evaluate the utility of the proposed biomedical entity representations, we use them 
for the tasks of biomedical entity label classification (ELC) and named entity disambiguation (NED). We highlight that these models perform well even without fine-tuning, which is critical in low- or zero-supervision scenarios.

\subsection{NED on Biomedical Wikipedia articles}
The NED task connects entity mentions in text with real world entities in a knowledge base by disambiguating the true entity from a list of candidates. 
We consider the local resolution setting in which each instance features a single entity mention span in the input text and several possible candidates with corresponding descriptions (e.g., the first paragraph of their Wikipedia article). 

\renewcommand{\arraystretch}{1}
\begin{table}[!t]
\centering
\small
\setlength{\tabcolsep}{4pt}
\begin{tabular}{lllll}
\toprule
\multicolumn{1}{l}{Dataset} & 
\multicolumn{1}{c}{Mentions} & 
\multicolumn{1}{c}{Abstracts} & 
\multicolumn{1}{c}{Type} & 
\multicolumn{1}{c}{Sets} \\
\midrule
MedMentions  & 350K &   4.3K-PubMed   & gold  & no              \\
BIERs*   & 37M  & 460K-PubMed   & silver & no              \\
ClinWikiNED* & 10K & 35K-Wiki & silver & yes    \\
\bottomrule
\end{tabular}
	\caption{Comparison of BioMed Linked Datasets. BIERs and ClinicalWikiNED datasets described in Section 3 and 4.1 respectivtely}
	\label{tab:BioLinked}
\end{table}

\paragraph{NED dataset construction}
 
While there exist multiple biomedical named entity recognition and linking datasets \cite{DBLP:journals/corr/abs-1902-09476,basaldella2020cometa}, we did not find much in the way of publicly available biomedical NED corpora, and we therefore constructed a new dataset, which we will release for use by other researchers.
The dataset is based on the set of Wikipages used by \citet{rosenthal-etal-2019-leveraging}, as relevant medical literature which consists of 34,692 medically relevant articles under the ‘Clinical Medicine’ category \footnote{https://en.wikipedia.org/wiki/Category:Clinical\_medicine}. We used SLING\footnote{Based on a June 1, 2020 dump of English wikipedia.} to process these articles and were able to retrieve around 1.5 million training examples (mention, context, [categories]) from them.  

After obtaining these examples for each entity mention we used the CrossWikis dictionary \citep{Spitkovsky_12} to try to gather between 3 to 5 challenging candidate entities for the example. 
This range in terms of number of candidates was selected because we wanted to include salient biomedical terms that are difficult to disambiguate; setting a higher number of potential candidates for use with CrossWikis largely gives general and short ``popular'' candidates (i.e., those that appear often in Wikipedia). 
This behavior makes sense since many biomedical terms are quite specific and usually only have a few high quality alternative candidates to select from.
Additionally, we filter out redirect pages and pages that no longer match the wiki version used to create CrossWikis. 

This candidate generation and data content acquisition step filters out considerably the number of available examples. We additionally subsample the dataset to reduce the instances where the ``popular'' candidate is the correct entity so as to make the task more difficult and to allow for more rare entities to appear in our set.  After all the filtering, our ClinicalWikiNED dataset consists of a train/dev/test split of size 5332, 3730, and 800 respectively. Table \ref{tab:BioLinked} shows a comparison of the two datasets introduced in this paper with that of one of the largest publicly available linked biomedical datasets\cite{murty-etal-2018-hierarchical}. 

\paragraph{Using BIERs for NED} Using the BIER architecture, we first train a separate {\tt WikiDescription} model that takes as input a wikipage title as its mention, its first paragraph as the context, and outputs a sparse embedding that predicts the page's categories. As training data, we use any Wikipedia page that contains categories in our biomedical entity type system. We use 2.5 million such (title, descriptions, [categories]) as our training data, and we check for model convergence on a small development set. This model is used so that candidate embedding dimensions will align with our BIER mention-context embeddings.

For each mention $m$ and context $s$ in the test set, we use a BIER model to induce a sparse representation $t$. 
We then go through each candidate $c_i$ for the current test example and use the {\tt WikiDescription} model to retrieve the candidate's sparse output embedding $t_{c_i}$. 
Finally, we compute both the cosine similarity and dot product of $t$ with each candidate $t_{c_i}$ and predict the candidate ${c_i}$ that achieves the highest score for each metric as the true one. 

\renewcommand{\arraystretch}{1}
\begin{table}[!t]
\centering
\small
\setlength{\tabcolsep}{4pt}
\begin{tabular}{l c c}
\toprule
\multicolumn{1}{l}{} & \multicolumn{2}{c}{Test Acc.}\\
\cmidrule(r){2-3}
\multicolumn{1}{c}{Model} & \multicolumn{1}{c}{Dot Prod} & \multicolumn{1}{c}{Cosine Sim} \\
\midrule
BIER-PubMedBERT (ours) & 80.1  & \textbf{84.0} \\  
BIER-SciBERT (ours) & 76.4 & 77.3 \\  
BIER-BioBERT (ours) & 71.9 & 75.9 \\  
\midrule
\citet{onoe-durrett-2020-interpretable} & 63.6 & 69.8 \\  

Popular Prior   &  73.9 & -\\
PubMedBERT \citep{gu2020domainspecific}  &  77.6  & - \\  
SciBERT \citep{Beltagy-etal-2019-scibert} &  77.4 & - \\
BioBERT \citep{10.1093/bioinformatics/btz682} & 77.9 & - \\
\bottomrule
\end{tabular}
	\caption{BIER zero shot test results vs Logistic Regression Baselines trained on task data for NED task}
	\label{tab:NEDres}
\end{table}

\paragraph{Baseline model for NED} 
We use the EntEval \cite{Mingda_Chen_19} framework  for our experiments and train a logistic regression classifier using a feature vector composed of the mention-context embedding $x_1$ and current candidate wiki description embedding $x_2$ from the set of candidates $C_m$ as a concatenation of $x_1$, $x_2$, element-wise product, and absolute difference: $[x_1, x_2, x_1 \odot x_2, |x_1 - x_2|]$.  Both $x_1$ and $x_2$ are obtained via BERT based models. Training minimizes binary log loss using all negative examples. At test time, inference combines this classifier result with the prior probability of how frequently candidates occur in Wikipedia as follows:
$\textit{arg max}_{c\in C_m} [p_{prior}(c) + p_{classifier}(c)]$ to obtain the final candidate prediction. Directly using the most likely prior as predictions yields an accuracy of 73.9\%.  We emphasize that these baselines are fine-tuned on the task data while the BIER models only do inference on the test set.

\paragraph{Results}
Table ~\ref{tab:NEDres} shows the results of the NED experiments. The biomedical BIER model affords improvements over the prior general domain interpretable model \cite{onoe-durrett-2020-interpretable}, showing that the biomedical type system and training is beneficial for this type of task.
In addition, the BIER models outperform the baselines without fine-tuning on the training data.

\subsection{ELC on Cancer Genetics data}

For our entity label classification task we use the Cancer Genetics dataset \cite{pyysalo-etal-2013-overview} which consists of 10,935 training, 3,634 dev, and 6,955 test examples from 300, 100, and 200 unique PubMed articles, respectively.\footnote{In our experiments we combine the train and dev sets into a single training set.} 
Given an article title and abstract, mention, and the corresponding entity label, the objective is to predict this label from 16 available coarse labels (see Table~\ref{tab:ELlabeldist} in the Appendix for label distribution information).

To assess how well the learned BIER representations fare against comparable baselines, we perform a simple nearest neighbor classification technique using the proposed BIER model variants, the general domain model from \citet{onoe-durrett-2020-interpretable}, and non-BIER fine-tuned pre-trained language models as standalone encoders.

We first induce dense embeddings for all training examples by passing the mention $m$ and context $s$ through the encoders as {\tt[CLS]} $m$  {\tt[SEP]} $s$ {\tt[SEP]}, and we store the resultant contextualized {\tt[CLS]} embedding $\mathbf{h}_{\texttt{[CLS]}}$ as our dense embedding. 
For the BIER and \citet{onoe-durrett-2020-interpretable} models we also save the final sparse entity embedding $\mathbf{t}$. 

We iterate over the test examples and similarly induce dense representations for these $\mathbf{h}^{test}_{\texttt{[CLS]}}$ and (if applicable) sparse representations $\mathbf{t}^{test}$.
We find their nearest neighbor (under either $\ell$2 distance or dot product similarity) from the saved training set of embeddings, and use its label as the prediction. 
We use the FAISS semantic indexer \cite{JDH17} for storing embeddings and finding nearest neighbors quickly.
We are interested in evaluating the off-the-shelf utility of learned representations, and, as such, we do not train or fine-tune the models in any of these cases; rather, training examples are used only for nearest neighbor retrieval.

That said, for completeness we also performed additional experiments in which we do fine-tune models on the task data, with varying amounts of supervision; we are interested especially in low-supervision settings. 
For the fine-tuning experiment, we add a linear layer on top of the best performing BIER and baseline models, using cross entropy loss as our objective and fine-tuning them for 4 epochs on the training data before performing inference. For the low supervision regime experiment, we show how the best nearest neighbor and fine-tuned models perform when given $K$ examples per class for $K \in [5,10,25,50,75, 100,200]$

\renewcommand{\arraystretch}{1}
\begin{table}[t]
\centering
\small
\setlength{\tabcolsep}{4pt}
\begin{tabular}{lcccc} 
\toprule
\multicolumn{1}{l}{} & \multicolumn{4}{c}{Test Acc.}\\
\cmidrule(r){2-5}
\multicolumn{1}{l}{} &  \multicolumn{2}{c}{L2 Dist} & \multicolumn{2}{c}{Dot Prod}\\
\cmidrule(r){2-3} \cmidrule(r){4-5}
\multicolumn{1}{c}{Model} & \multicolumn{1}{c}{Dense} & \multicolumn{1}{c}{Sparse} & \multicolumn{1}{c}{Dense} & \multicolumn{1}{c}{Sparse}\\
\midrule
BIER-PubMedBERT & 85.5 & 86.8 & \textbf{88.2} & \textbf{87.5}  \\
BIER-SciBERT    & 70.8 & 77.0 & 72.8 & 76.8 \\
BIER-BioBERT    & 83.4 & 85.9 & 85.6 & 86.8 \\
\midrule
\citet{onoe-durrett-2020-interpretable} &  63.9  & 55.1 & 60.0 &  59.9 \\
PubMedBERT      & 77.3 & - & 69.3 & - \\
SciBERT         & 74.4 & - & 75.2 & - \\
BioBERT         & 67.6 & - & 59.6 & - \\
\bottomrule
\end{tabular}
	\caption{Test accuracy on Cancer Genetics data using a nearest neighbor classifier (k=1) without fine-tuning based on sparse output or intermediate dense embeddings using L2 or Dot Product distance metrics. }
	\label{tab:ELres}
\end{table}

\paragraph{Results}
Table ~\ref{tab:ELres} shows the results for our first experiment, in which we use untuned representations.
We observe that the baseline language model encodings all perform worse than the proposed BIER sparse and dense models, with the exception of SciBERT, which fares better than the sparse BIER model based on SciBERT. 
Additionally, we see that BERT and \citet{onoe-durrett-2020-interpretable} (which is based on BERT) both perform poorly in this biomedical task compared to the other baselines. 

Importantly, we notice that the sparse interpretable embedding results for our top performing models (both BIER-PubMedBERT and BIER-BioBERT) perform near the level of their dense, non-interpretable counterparts.  In the next section we will look at some illustrative test examples cases along with a simple technique to leverage both the dense and sparse embeddings that a BIER model can give to improve performance on the task and gain insight into where the entity type model and system may be underperforming.

Table ~\ref{tab:ELres-finetune} shows the results of our fine-tuning experiment.  Freezing the model and allowing only the classification layer to learn weights doesn't allow enough capacity for either case, while fully fine-tuning both models gives improved performance in both models.  However because the BIER model is no longer tied in, 
the interpretability component of our representations is eliminated, a limitation left for future work.    

\begin{figure}
    \includegraphics[width=1.0\linewidth]{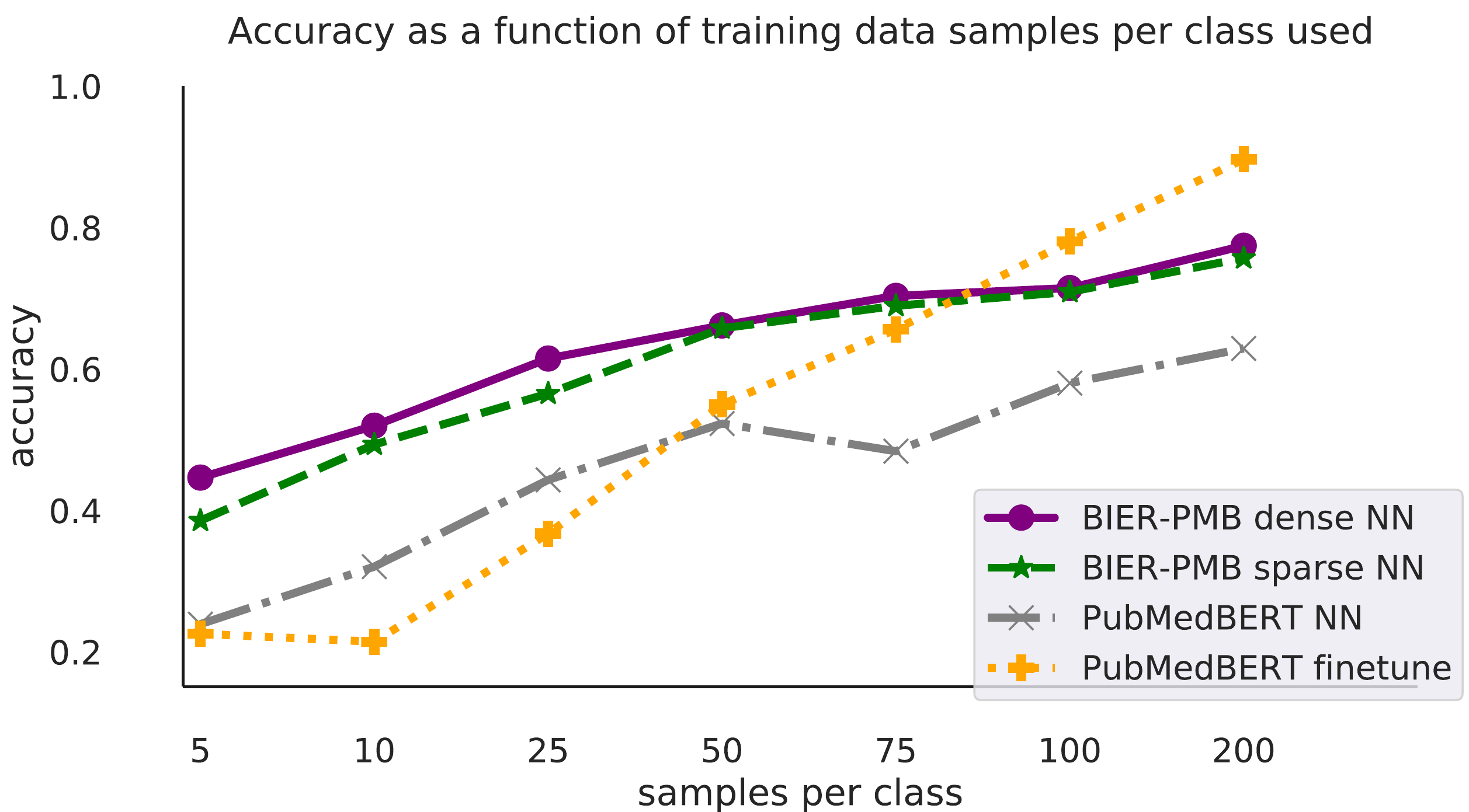}
    \caption{Results for the entity label classification task under varying amounts of supervision.}
    \label{fig:lowdata}
\end{figure}

Figure ~\ref{fig:lowdata} shows BIER-PubMedBERT performs better than the fine-tuned and non-interpretable PubMedBERT model when there are fewer than 100 examples per class ( which is the case for 6 out of the 16 test classes in the dataset as seen in table ~\ref{tab:ELlabeldist} in the appendix).

\renewcommand{\arraystretch}{1}
\begin{table}[t]
\centering
\small
\setlength{\tabcolsep}{4pt}
\begin{tabular}{lcc} 
\toprule
\multicolumn{1}{l}{} & \multicolumn{2}{c}{Test Acc.}\\
\cmidrule(r){2-3}
\multicolumn{1}{c}{Model} & \multicolumn{1}{c}{Frozen Model} & \multicolumn{1}{c}{Fine-Tuned} \\
\midrule
BIER-PubMedBERT (ours) & 68.0  & 96.0\\
PubMedBERT & 36.2 & 96.1\\

\bottomrule
\end{tabular}
	\caption{Test results on Cancer Genetics task with fine tuning on all data whether freezing the model or not.}
	\label{tab:ELres-finetune}
\end{table}

\section{Debugging with BIERs}

One of the claimed advantages of BIERs is their ability to facilitate model debugging.
In this section we provide illustrative examples where the interpretability of the underlying representations offers insights into model behavior and suggests avenues for improvements.

\paragraph{Entity Type and Mention Analysis}

We illustrate the debugging strategy proposed in the context of entity label classification. 
Recall that this entails inspecting test examples for which the dense model yields a correct prediction, while the sparse variant does not (implying that the former somehow better represents the instance).  
We can inspect these cases to understand what entity types are leading to such behavior.  
Appendix Table \ref{tab:ned-de-not-sp-1} and \ref{tab:ned-de-not-sp-2} enumerate such mentions and their most probable types. 
We note the inclusion of many people's names (e.g., ``Anthony Campbell", ``Tony Walsh") which have been assigned at least some incorrect types in their sparse representations. 
This highlights a general failure mode of the model: It is assigning incorrect types to person names, which may be causing downstream prediction errors. 
This is actionable information, as we could remedy the issue via rules, additional, targeted supervision or by down weighting probabilities given to common erroneous types for these mentions.

To better characterize entity type errors, we gather the set of the 20 most probable entity types for all mentions incorrectly predicted by the BIER sparse model and sort types by frequency. We do the same for those predicted correctly. 
The resulting two lists share many of the same top popular types, but looking at relative rankings and only displaying those 
that are comparatively far apart\footnote{We chose to highlight entity types that are farther than 50 rankings apart to have a small set to display.}  reveals some interesting results. 
Tables \ref{tab:ELwrongtypes} and \ref{tab:ELrighttypes} report entity types correlated with correct and incorrect predictions, respectively.
We emphasize this type of analysis is only possible due to the interpretable nature of the proposed BIER embeddings. 

As a final illustrative debugging example, we consider a test example mention ``thyroid carcinomas” with label ``Cancer”, along with the predictions made by the sparse model, ``thyroid” with the incorrect label ``Organ", and the dense model,``esophageal carcinoma” with the correct label ``Cancer”.
We also retrieve the first correct prediction from the nearest neighbors of the sparse model embedding ``medullary thyroid carcinoma” which we refer to as the \emph{counterfactual} sparse prediction.\footnote{Had the mention under consideration instead mapped to this sparse representation, the prediction would have been different, and correct.}
We take the dot product of the mention-context embedding with these three prediction’s embeddings and inspect the top types which lead to their selection in Figure \ref{fig:bier_analysis} in Appendix C. 
Both the incorrect sparse and correct counterfactual sparse predictions, at the surface level are quite similar to the test mention, but have lower scores for the entity type ‘thyroid cancer’ compared with the dense prediction which gives the correct label, but is semantically less similar to the test mention than the counterfactual sparse prediction. Additionally, the noisy type ``rtt” erroneously plays more of a role in the sparse model predictions as well.

\paragraph{Diagnosing task results} In analyzing errors made by the highest performing BIER dense and sparse nearest neighbor models for the entity label classification task, we noticed that while there was high concurrence for correct predictions (i.e., of the 88\% true predictions made by the dense model overall, the sparse model agreed with the prediction 95\% of the time), the cases where the model predictions disagreed, but where one of them still predicted the true label, were quite varied. 
 In other words, the sparse model gave many correct results on test cases when the dense model gave incorrect ones and vice versa. Applying the diagnostic technique from Section 3, we see the classifier's overall performance could have improved from 88.2 to 91.9 had the model known when to utilize its intermediate dense representation over its sparse output.

Similarly we applied the diagnostic technique to the NED task and leave more details in Appendix B.  Incorporating mentions that the dense dot product BIER model handles better than the cosine similarity based sparse one does would have given an improvement from our prior accuracy of 84.0 to 91.7.  Table \ref{tab:combined} shows the possible improvement in task accuracies for both tasks.

\renewcommand{\arraystretch}{1}
\begin{table}[t]
\centering
\small
\setlength{\tabcolsep}{4pt}
\begin{tabular}{lcccc} 
\toprule
\multicolumn{1}{l}{} & \multicolumn{3}{c}{Test Acc.}\\
\cmidrule(r){2-4}
\multicolumn{1}{c}{Task} & \multicolumn{1}{c}{Dense} & \multicolumn{1}{c}{Sparse} & \multicolumn{1}{c}{Combined} & \multicolumn{1}{c}{$\Delta$} \\
\midrule
NED  & 84.0 & 81.0  & \textbf{91.7} & +7.7 \\
ELC   & 87.5 & 88.2  & \textbf{91.9} & +3.7 \\
\bottomrule
\end{tabular}
	\caption{Results for both tasks showing improvements that could have been achieved by combining intermediate dense and interpretable sparse output embeddings generated by the same BIER-PubMedBERT model. 
	}
	\label{tab:combined}
\end{table}

\section{Related Work}

In this work we have introduced a predefined fine-grained biomedical type system comprising 68k types, explicitly tied to PubMed. 
Instead of using a fixed type system, \citet{Jonathan_Raiman_18} seek to dynamically learn a 100 dimensional type system from a much larger general domain type system in order to optimally disambiguate entities.

Aside from work on biomedical NLP and entities specifically, there exists a line of work on interpretable word embeddings \citep{Anant_Subramanian_17, Faruqui2015SparseOW}. A common approach here is to 
identify the groups of words most associated with vector components globally, somewhat akin to topic models. This differs from our approach, which is based on an external type system and provides immediate, instance-level interpretable probabilities for each entity type.  \citet{hu-etal-2020-transformation} proposes transforming dense to sparse representations 
independent of entity typing.

Another related line of work tests a models’ ability to induce syntactic or type information by the measuring accuracy of a probe \cite{Matthew_Peters_18,John_Hewitt_19_a, John_Hewitt_19_b}.  There is significant uncertainty about how to calibrate such post-hoc probing results \cite{Elena_Voita_20} whereas our model’s representations are directly interpretable.

While many interesting biomedical entity representation and linking task oriented works \cite{murty-etal-2018-hierarchical, vashishth2020medtype, mondal-etal-2019-medical, sung-etal-2020-biomedical, liu2020selfalignment} leverage PubMed or UMLS for semantic type, entity synonym, or self alignment purposes, our work is the first to incorporate interpretable embeddings that are linked to a biomedical entity type system.

\section{Conclusions}
We have introduced a new biomedical entity typing system and training set from a large corpus of biomedical texts.
We will release this dataset, which comprises 37 million derived triples. Exploiting this data, we proposed \emph{Biomedical Interpretable Entity Representations} (BIERs), in which dimensions correspond to fine-grained entity types, and values are predicted probabilities that a given entity is of the corresponding type. 

Using two downstream biomedical tasks, we showed that BIER representations yield predictive performance that is competitive with dense (uninterpretable) representations, and that such representations are particularly beneficial in zero-shot or low-supervision settings. 
We also demonstrated that BIER representations can facilitate meaningful model debugging both at the mention and entity type level. 

\section*{Acknowledgements}

This material is based upon work supported in part by the National Science Foundation (under Grants No. NSF 1901117 and NSF 1750978) and in part by a grant from the Office of Naval Research (ONR N00014-17-1-2143). The project was conducted under the auspices of the IBM Science for Social Good initiative.

\section*{Ethical Considerations}

NLP models are increasingly used in biomedicine, where some applications can be quite high-stakes. 
Establishing trust in such models is therefore paramount; unfortunately, deep neural networks tend to be opaque in their operations, potentially precluding their use in certain areas of biomedicine where they might otherwise be beneficial.
This work is a step towards more transparent models for biomedical NLP.

\bibliographystyle{acl_natbib}
\bibliography{acl2021}

\newpage
\clearpage
\appendix

\section{BIER system level specifics}\label{app:concepts}
To better illustrate the process of which mentions are retained during our filtering process, Table ~\ref{tab:concepts} shows the 6 concepts associated with an example mention of "phase II clinical trial" found in a PubMed article.  We see all 6 concepts score higher than our minimum threshold and we use the two highest scoring matches that are within 2 points of each other: CUIDs C0282460 and C1096779. the former C0282460 has a WikiPedia data item Q7180990 that corresponds to the page ``wiki/Phases\_of\_clinical\_research" whose associated categories are ``Clinical research", ``Design of experiments", ``Life sciences", ``industry". 
The second result C1096779 has no direct WikiPedia match and the results we get from SLING include ``Clinical trial", ``Scientific control", ``Medicine", ``Topical medication", ``Observational study", ``Literature".  Hence for this mention and context from a PubMed abstract, we are able to extract a (mention, context, list of types) triple of the form (``phase II clinical trial", context, [``Clinical research", ``Design of experiments", ``Life sciences" industry", ``Clinical trial", ``Scientific control", ``Medicine" .... ]].

\begin{table}[h]
\centering
\small
\setlength{\tabcolsep}{4pt}
\begin{tabular}{l|l|l|l}
\toprule
\multicolumn{1}{c|}{\textbf{CUID}} &  \multicolumn{1}{c|}{\textbf{Concept Name}} & \multicolumn{1}{c|}{\textbf{Score}} & \multicolumn{1}{c}{\textbf{DBPedia}} \\
\midrule
C0282460 &  Phase 2 Clinical Trials &  0.9999 &  Q7180990 \\
C1096779 &  Clinical Trial, Phase II & 0.9999 &  none \\
C0282461 &  Phase 3 Clinical Trials &  0.9496 &  Q7180990 \\
C0920321 &  Phase I Clinical Trials &  0.8707 &  Q7180990 \\
C1096780 &  Clinical Trial, Phase III &  0.8635 &  none \\
C0282462 &  Phase 4 Clinical Trials &  0.8208 &  Q7180990\\
\bottomrule
\end{tabular}
  
  \caption{Using an NER tagger we find 6 associated concepts in UMLS for the mention ``phase II clinical trial" in a context sentence ``Unraveling the molecular mechanism of BNC105, a phase II clinical trial vascular disrupting agent, provides insights into drug design."}
  
  \label{tab:concepts}
\end{table}


\begin{figure}[t]
    \centering
    \includegraphics[width=1.0\linewidth]{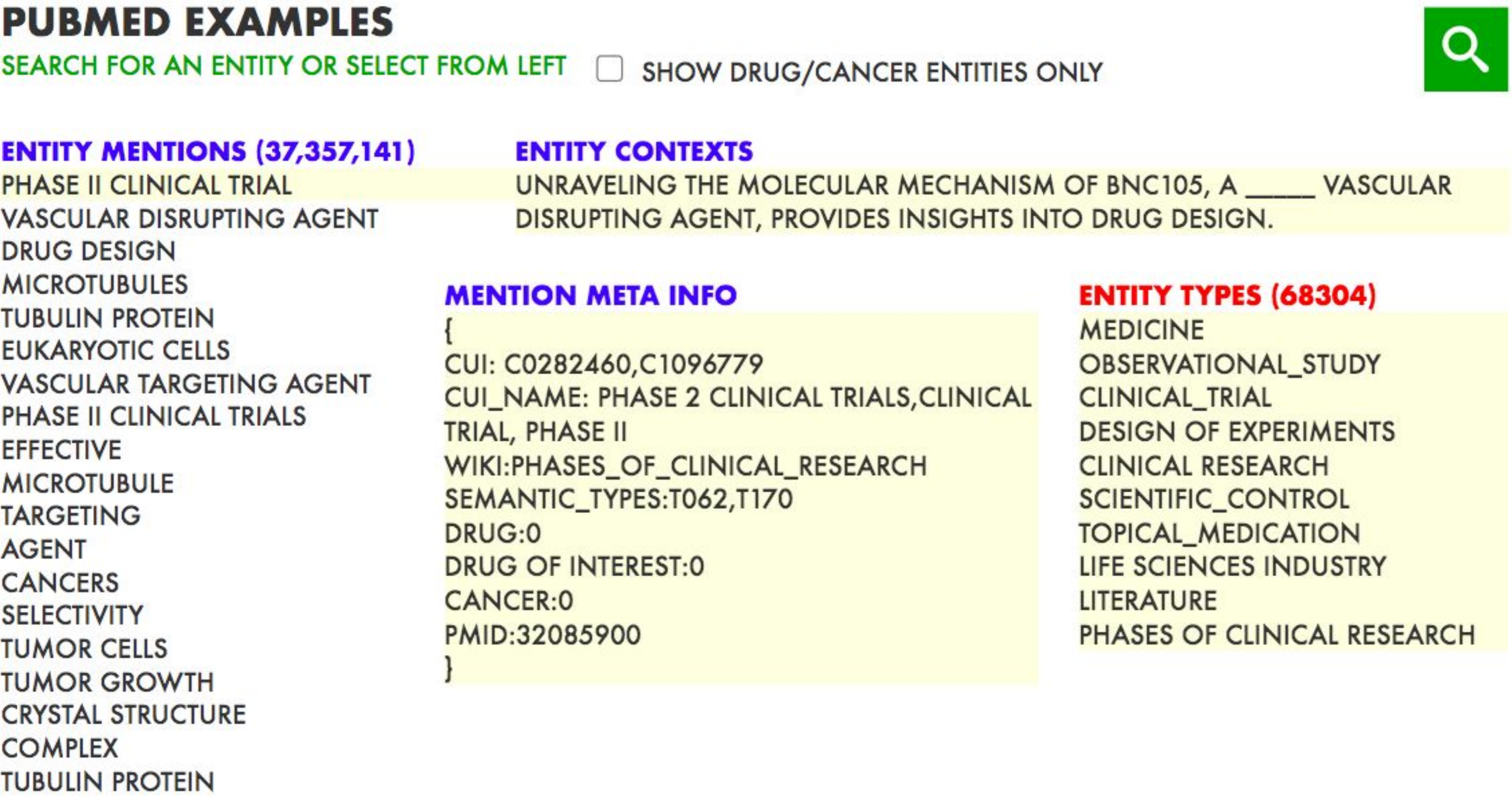}
    \caption{Example from derived Biomedical Dataset}
    \label{fig:big_pub}
\end{figure}



\renewcommand{\arraystretch}{1}
\begin{table}[!t]
\centering
\small
\setlength{\tabcolsep}{4pt}
\begin{tabular}{l|c|c} 
\toprule
\multicolumn{1}{c|}{\makecell{Label}} & \multicolumn{1}{c|}{\makecell{Train + Dev\\Set \% ( raw )}} & \multicolumn{1}{c}{\makecell{Test Set\\ \% ( raw )}} \\
\midrule
Gene\_or\_gene\_product           & 36.98 ( 5388 )                    & 36.23 ( 2520 )                \\
Cell                              & 17.32 ( 2524 )                  & 15.15 ( 1054 )                \\
Cancer                            & 11.52 ( 1679 )                    & 13.30 ( 925 )                 \\
Simple\_chemical                  & 10.59 ( 1543 )                   & 10.45 ( 727 )                \\
Organism                          & 8.63 ( 1258 )                    & 7.81 ( 543 )                \\
Multi-tissue\_structure           & 3.80 ( 554 )                     & 4.36  ( 303 )              \\
Tissue                            & 2.77 ( 403 )                   & 2.73 ( 190 )                \\
Cellular\_component               & 2.67 ( 389 )                    & 2.59 (180 )                \\
Organ                             & 1.82 ( 265 )                   & 2.24 ( 156 )               \\
Organism\_substance               & 1.24 ( 181 )                  & 1.47 ( 102 )                \\
Pathological\_formation           & 0.96 ( 140 )                    & 1.28 ( 89 )                \\
Amino\_acid                       & 0.50 ( 73 )                     & 0.89 ( 62 )                \\
\makecell{Immaterial\_anatomical\\\_entity}    & 0.49 ( 71 )                    & 0.45 ( 31 )                \\
Organism\_subdivision             & 0.40 ( 59 )                     & 0.56 ( 39 )                \\
Anatomical\_system                & 0.16 ( 24 )                    & 0.24 ( 17 )                \\
\makecell{Developing\_anatomical\\\_structure} & 0.12 ( 18 )                    & 0.24 ( 17 )                \\
\bottomrule
\end{tabular}
	\caption{Cancer Genetics Dataset Label Distribution}
	\label{tab:ELlabeldist}
\end{table}

\section{NED diagnostic details}\label{app:ned_diagnostic}
For the NED task we used the BIER's sparse embeddings of test mentions in their contexts and took cosine similarity with a separate BIER model's sparse embeddings of candidate wiki descriptions to make our predictions. To use the diagnostic technique we first get task predictions using the dense embeddings from the BIER models which gives results of 81 and 79.25 percent test accuracy using dot product and cosine similarity respectively.  Although the prior sparse cosine similarity BIER model in this case gave a higher 84.0 percent test accuracy, using the diagnostic technique in this case by incorporating mentions the dense dot product BIER model handles better would have given an improvement in accuracy from 84.0 to 91.65.


\begin{figure}[t]
    \centering
    \includegraphics[width=0.8\linewidth]{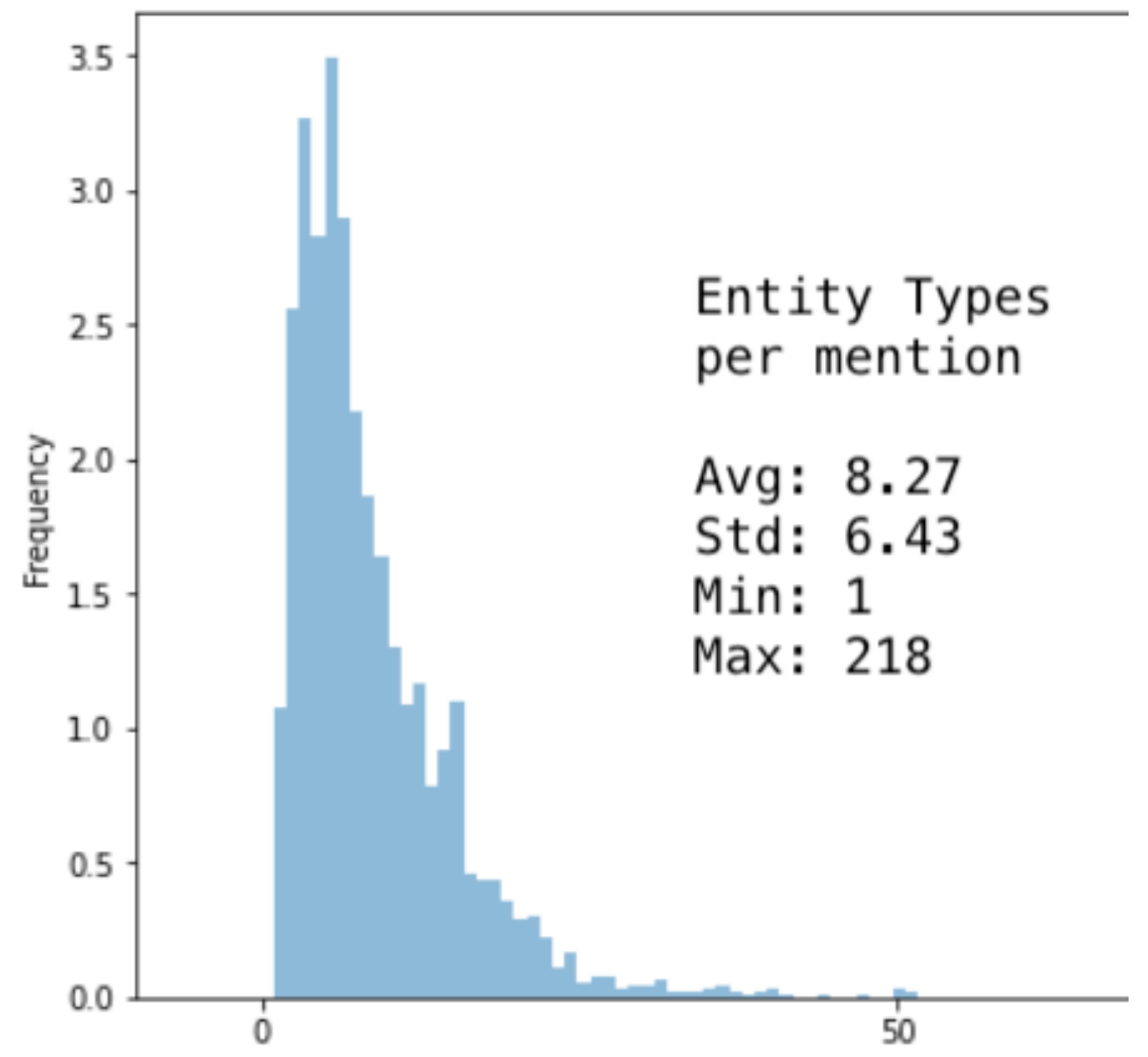}
    \caption{Entity Types per mention on Training set for BIER}
    \label{fig:bier_hist}
\end{figure}


\renewcommand{\arraystretch}{1}
\begin{table*}[t]
\centering
\small
\setlength{\tabcolsep}{4pt}
\begin{tabular}{l|l|l|l}
\toprule
\multicolumn{1}{c|}{1-25} & \multicolumn{1}{c|}{26-50} & \multicolumn{1}{c|}{51-75} & \multicolumn{1}{c}{76-100}                    \\ 
\midrule
term                            & w.h.o. essential medicines & test\_(assessment)               & causality                 \\
ingredient                      & psychosis                                     & drug discovery                   & hydroxyl                  \\
disease                         & scientific\_method                            & receptor\_(biochemistry)         & adverse\_effect           \\
cell\_(biology)                 & oncology                                      & observational\_study             & diagnosis                 \\
rtt                             & enzyme                                        & immunology                       & physiology                \\
protein                         & human\_body                                   & molecular biology                & chemotherapy              \\
gene                            & psychotherapy                                 & abnormality\_(behavior)          & hepatotoxins              \\
human                           & medicine                                      & radioactive\_decay               & molecule                  \\
neoplasm                        & grammatical\_modifier                         & derivative\_(chemistry)          & phenotype                 \\
cancer                          & tissue\_(biology)                             & chemistry                        & cell biology              \\
therapy                         & treatment\_and\_control\_groups               & health policy                    & concepts in metaphysics   \\
medical terminology             & scientific method                             & amine                            & concepts in epistemology  \\
measurement                     & coagulation                                   & peptide                          & apoptosis                 \\
patient                         & chemical\_reaction                            & pharmaceutical sciences          & procedural\_law           \\
chemical\_compound              & philosophy of science                         & antigen                          & science                   \\
surgery                         & calcium\_in\_biology                          & biology                          & genetic\_code             \\
nitrous\_oxide                  & enzyme\_inhibitor                             & algorithm                        & empiricism                \\
pharmaceutical\_drug            & medicinal chemistry                           & texas                            & family                    \\
acid                            & research                                      & mental\_disorder                 & thailand                  \\
articles containing video clips & metabolism                                    & statistical\_hypothesis\_testing & liver                     \\
malignancy                      & taxonomy\_(biology)                           & catalysis                        & medical mnemonics         \\
time                            & cell\_growth                                  & allele                           & dosage\_form              \\
prothrombin\_time               & blood                                         & methyl\_group                    & immune system             \\
cognition                       & syndrome                                      & infectious causes of cancer      & amino\_acid               \\
drug                            & sewage\_treatment                             & database                         & beta\_sheet       \\       
\bottomrule
\end{tabular}
	\caption{Top 100 most frequent types from Biomedical Entity Type System}
	\label{tab:types}
\end{table*}

\renewcommand{\arraystretch}{1}
\begin{table*}[h]
\centering
\small
\setlength{\tabcolsep}{4pt}
\begin{minipage}{0.46\linewidth}
\begin{tabular}{c|l|c|c} 
\toprule
\multicolumn{1}{c|}{\makecell{Incorrect\\rank}} & \multicolumn{1}{c|}{Entity Type} & \multicolumn{1}{c|}{\makecell{Correct\\rank}} & \multicolumn{1}{c}{\makecell{Relative\\difference}} \\
\midrule
20 & tongue & 76  &  56 \\
24 & anatomy & 160  &  136 \\
29 & protein\_domain & 112  &  83 \\
34 & organ\_(anatomy) & 107  &  73 \\
35 & gland & 205  &  170 \\
38 & phosphatase & 140  &  102 \\
43 & surgery & 120  &  77 \\
46 & circulatory\_system & 293  &  247 \\
50 & squamous-cell\_carcin & 111  &  61 \\
51 & nephron & 142  &  91 \\
60 & anatomical\_terms & 169  &  109 \\
61 & kidney & 284  &  223 \\
62 & cancer\_cell & 213  &  151 \\
70 & activator\_(genetics) & 179  &  109 \\
71 & drug & 192  &  121 \\
74 & breast\_cancer & 127  &  53 \\
75 & locus\_(genetics) & 206  &  131 \\
77 & cancer\_staging & 256  &  179 \\
79 & signal transduction & 233  &  154 \\
81 & multiprotein\_complex & 132  &  51 \\
82 & endometrium & 200  &  118 \\
83 & mouth & 200  &  117 \\
84 & cell anatomy & 272  &  188 \\
90 & molecular biology & 200  &  110 \\
93 & rare cancers & 200  &  107 \\
95 & website & 161  &  66 \\
96 & cell\_cycle & 200  &  104 \\
97 & gene expression & 178  &  81 \\
98 & hydroxyl & 221  &  123 \\
99 & oral\_sex & 200  &  101 \\
\bottomrule
\end{tabular}
	\caption{Entity Types more associated with erroneous predictions}
	\label{tab:ELwrongtypes}
\end{minipage}
\hspace{24pt}
\begin{minipage}{0.46\linewidth}
\begin{tabular}{c|l|c|c} 
\toprule
\multicolumn{1}{c|}{\makecell{Incorrect\\rank}} & \multicolumn{1}{c|}{Entity Type} & \multicolumn{1}{c|}{\makecell{Correct\\rank}} & \multicolumn{1}{c}{\makecell{Rel\\diff}} \\
\midrule
23 & syndrome & 244 & 221 \\
34 & abnormality\_(behavior) & 270 & 236 \\
35 & elementary\_particle & 128 & 93 \\
42 & apoptosis & 200 & 158 \\
51 & congenital\_disorder & 200 & 149 \\
53 & transformation\_(genetics) & 275 & 222 \\
54 & measurement & 109 & 55 \\
55 & human cells & 147 & 92 \\
56 & immune system disorders & 200 & 144 \\
57 & paraneoplastic syndromes & 200 & 143 \\
58 & code & 154 & 96 \\
59 & battery\_(electricity) & 200 & 141 \\
61 & virus & 222 & 161 \\
63 & chemistry & 281 & 218 \\
66 & calcium\_in\_biology & 209 & 143 \\
71 & thymus & 200 & 129 \\
72 & medical terminology & 190 & 118 \\
73 & cell biology & 297 & 224 \\
74 & recombinant\_dna & 10 & 64 \\
76 & tongue & 20 & 56 \\
79 & protein\_kinase & 164 & 85 \\
80 & drama & 200 & 120 \\
85 & tumor\_suppressor\_gene & 14 & 71 \\
86 & patient & 200 & 114 \\
87 & specialty\_(medicine) & 234 & 147 \\
90 & growth\_hormone & 200 & 110 \\
91 & taxonomy\_(biology) & 238 & 147 \\
93 & t cells & 228 & 135 \\
94 & childhood & 200 & 106 \\
95 & aging-related proteins & 200 & 105 \\
96 & network\_affiliate & 200 & 104 \\
97 & blood tests & 200 & 103 \\
98 & protein\_a & 200 & 102 \\
\bottomrule
\end{tabular}
	\caption{Entity Types more associated with correct predictions}
	\label{tab:ELrighttypes}
\end{minipage}
\end{table*}


\clearpage

\renewcommand{\arraystretch}{1}
\begin{table*}[t]
\centering
\small
\setlength{\tabcolsep}{4pt}
\begin{tabular}{c|l} 
\toprule
\multicolumn{1}{c|}{\textbf{Mention}} & \multicolumn{1}{c}{\textbf{Sparse embedding top types that do worse than dense counterparts}} \\
\midrule
                           & albinism, disease, animal coat colors, heredity, dermatologic terminology,\\
\multirow{-2}{*}{albinism} & articles containing video clips, hair, skin, pigment, nitrous\_oxide \\
\rowcolor{Gray} & tooth, lung anatomy, mouth, pulmonary\_alveolus, human mouth anatomy, dental\_caries, \\
\rowcolor{Gray} \multirow{-2}{*}{alveolar ridge} & periodontology, parts of tooth, mandible, leaf \\

 & surgery, anastomosis, evolutionary biology, digestive system, angiology, anatomy, lawsuit, \\
\multirow{-2}{*}{anastomosis} & combat, organ\_(anatomy), surgical\_anastomosis \\

\rowcolor{Gray} & carl\_linnaeus, ingredient, human, taxa named by carl linnaeus, flora of asia,\\
\rowcolor{Gray} & world health organization essential medicines, coordination\_complex, \\
\rowcolor{Gray} \multirow{-3}{*}{anthony campbell} & western european countries, extract, asteraceae genera \\

 & peripheral nervous system disorders, autonomic nervous system, disease, nervous\_system, \\
\multirow{-2}{*}{autonomic neuropathy} & peripheral\_neuropathy, functional\_group, heredity, mental\_disorder, nerve \\

\rowcolor{Gray} & skin conditions resulting from physical factors, lesion, fluid, frostbite, radiation health effects, \\
\rowcolor{Gray} \multirow{-2}{*}{bleb} & disease, source\_code, nitrous\_oxide, skin, hematology \\

 & cell\_(biology), cell\_culture, medical terminology, oncology, cancer, precancerous\_condition,\\
\multirow{-2}{*}{cancer cells} & large\_cell, human, protein, standard\_operating\_procedure \\

\rowcolor{Gray} & mitochondria, programmed cell death, death, cell\_(biology), cellular senescence, cognition, \\
\rowcolor{Gray} \multirow{-2}{*}{cell death} & apoptosis, tgf\_beta\_signaling\_pathway, nuclear\_receptor, survival\_rate \\

 & plasma\_cell, small\_intestine, mood\_disorder, b-cell\_lymphoma, lymphocytic leukemia, \\
\multirow{-2}{*}{\makecell{chronic lymphocytic\\leukemia}} & bioaccumulation, bone\_marrow, grading\_(tumors), lymphatic\_system, lymphoblast \\

\rowcolor{Gray} & ionizing radiation, assumption, units\_of\_measurement, comics by steve ditko, cell\_(biology),\\
\rowcolor{Gray} \multirow{-2}{*}{cosmological constant} & grammatical\_modifier, quantity, blood\_plasma, industrial gases, litre\\

 & organic reactions, gene expression, posttranslational modification, rna, transcription\_(genetics),\\
\multirow{-2}{*}{demethylation} & molecular genetics, epigenetics, therapy, demethylation, molecular biology \\

\rowcolor{Gray} & wine regions of south africa, suburbs of cape town, astronomical\_unit \\
\rowcolor{Gray} \multirow{-2}{*}{dissociation constant} & elementary\_particle, rat, medical terminology, gene, units\_of\_measurement, furans \\

 & endoscopy, bicycle, diagnostic gastroenterology, physical\_examination, gastroenterology, \\
\multirow{-2}{*}{endoscope} & microphone, video\_camera, israeli inventions, pencil, 21st-century inventions \\

\rowcolor{Gray} & finger, conditions of the skin appendages, articles containing video clips, disease, toe, \\
\rowcolor{Gray} \multirow{-2}{*}{fingering} & nitrous\_oxide, hand, keratin, reflex, fingers \\

 & female, causes of death, fly, metrorrhagia, disease, articles containing video clips,\\
\multirow{-2}{*}{flirting} & conditions of the skin appendages, etiology, dog, homology\_(biology) \\

\rowcolor{Gray} & tongue, anatomical\_terms\_of\_location, ganglion, mandible, midbrain, organ\_(anatomy), \\
\rowcolor{Gray} \multirow{-2}{*}{geniculate} & cell\_nucleus, cerebral\_cortex, lobe\_(anatomy), middle\_ear \\

 & kidney, tongue, connective\_tissue, epithelium, gene, cell\_membrane, nitrous\_oxide, \\
\multirow{-2}{*}{glomerulus} & organ\_(anatomy), nephrology, derivative\_(chemistry) \\
 
\rowcolor{Gray}  & hemoglobins, respiratory physiology, hemoglobin, geography, equilibrium chemistry, cancer,\\        
\rowcolor{Gray} \multirow{-2}{*}{guy davis} & race\_and\_ethnicity\_in\_the\_united\_states\_census, geographic\_coordinate\_system, texas\\

 & infant, infant feeding, child, milk, formula, dosage\_form, foods, breast milk,\\
\multirow{-2}{*}{infant formula}  & preterm\_birth, chemistry \\

\rowcolor{Gray} & bowel\_obstruction, human\_gastrointestinal\_tract, large\_intestine, invagination, disease, \\
\rowcolor{Gray} \multirow{-2}{*}{intussusception} & morphology\_(biology), colorectal\_cancer, nitrous\_oxide, deconstruction, thrombosis \\
  
 & isomerism, stereochemistry, metabolism, 1827 introductions, laboratory techniques,\\
\multirow{-2}{*}{isomerization} & transgender, chemistry, organic chemistry, isomerases, flora of california \\

\rowcolor{Gray} & ingredient, rtt, mesylate, abbvie inc. brands, anti-inflammatory, orphan drugs, acid,\\
\rowcolor{Gray} \multirow{-2}{*}{mescaline} & methyl\_group, carbamates, amine \\

 & epigenetics, posttranslational modification, amino\_acid, protein, methylation, acid,\\
\multirow{-2}{*}{methylation} & organic reactions, amine, antigen, ingredient \\

\bottomrule
\end{tabular}
\caption{NED examples where dense BIER embeddings outperforms sparse (interpretable) BIER representations. Mentions start with [A-M].}
\label{tab:ned-de-not-sp-1}
\end{table*}

\renewcommand{\arraystretch}{1}
\begin{table*}[t]
\centering
\small
\setlength{\tabcolsep}{4pt}
\begin{tabular}{c|l} 
\toprule
\multicolumn{1}{c|}{\textbf{Mention}} & \multicolumn{1}{c}{\textbf{Sparse embedding top types that do worse than dense counterparts}} \\
\midrule
            
n400 & antigen, cancer, gene, protein, units\_of\_measurement, ratio, allele, human, time, nucleolus \\
     
\rowcolor{Gray} & acne\_vulgaris, topical\_medication, ingredient, functional\_group, route\_of\_administration, peroxides,\\
\rowcolor{Gray} \multirow{-2}{*}{peroxides} & chemical\_reaction, glandular and epithelial neoplasia, functional groups, pharmaceutical\_drug \\
          
 & nutrition, fatty acids, acid, lipids, ester, protein, ingredient, food science, \\
\multirow{-2}{*}{\makecell{polyunsaturated\\fatty acids}} &  neuronal\_ceroid\_lipofuscinosis, lipid \\

\rowcolor{Gray} & endopeptidase, enzyme\_inhibitor, biosynthesis, protease, chemical\_compound, receptor\_antagonist,\\
\rowcolor{Gray} \multirow{-2}{*}{protease inhibitors} & peptide, moa, hiv, hiv-1\_protease \\

 & psychology, substance\_dependence, substance\_abuse, emotion, mental\_disorder, dependent territories,\\
\multirow{-2}{*}{\makecell{psychological\\dependence}} & governance of the british empire, mental and behavioural disorders, crown dependencies, british islands \\

\rowcolor{Gray} & abnormal psychology, abnormality\_(behavior), nitrous\_oxide, pathology, disease, behavioural sciences,\\
\rowcolor{Gray} \multirow{-2}{*}{psychopathy} & psychosis, mental and behavioural disorders, mental\_disorder, affect\_(psychology) \\
 
 & technology, natural\_resource, segmental\_resection, plant anatomy, plant physiology, surgical\_suture,\\
\multirow{-2}{*}{resection} & surgery, plant morphology, morphology\_(biology), amputation \\

\rowcolor{Gray} & epithelioid\_cell, etiology, chilblains, disease, nitrous\_oxide, kalashnikov derivatives, sarcoidosis,\\
\rowcolor{Gray} \multirow{-2}{*}{sarcoidosis} & organ\_(anatomy), 5.56×45mm nato assault rifles, carbines \\

 & tongue, ear\_canal, vestibular system, auditory system, eustachian\_tube, canal\_(anatomy), \\
\multirow{-2}{*}{semicircular canals} & auditory\_system, crystal\_structure, vestibulocochlear nerve, cranial\_cavity \\

\rowcolor{Gray} & sequence, bioinformatics, psychosis, psychoanalysis, dna, scientific method, nucleic\_acid\_sequence,\\
\rowcolor{Gray} \multirow{-2}{*}{sequence analysis} & physical\_examination, dna\_sequencing, algorithm \\

 & race\_and\_ethnicity\_in\_the\_united\_states\_census, adult, flora of asia, carl\_linnaeus, human,\\
\multirow{-2}{*}{tony walsh} & french-speaking countries,  flora of north america, hemoglobins, women, coagulation system \\

\rowcolor{Gray} & united\_states, united states federal executive departments, management, police, public\_health,\\
\rowcolor{Gray} & united\_states\_department\_of\_defense,  united states department of health and human services agencies,\\
\rowcolor{Gray} \multirow{-3}{*}{\makecell{united states\\department of\\agriculture}} & regulators of biotechnology products, 1889 establishments in the united states \\

 & ventricle\_(heart), zoning, ventricular\_system, ventricular system, brain, developmental neuroscience,\\
\multirow{-2}{*}{ventricular zone} & tongue, urban planning, anatomical\_terms\_of\_location, bone \\
                 
\rowcolor{Gray} & psychological testing, psychiatric assessment, connective/soft tissue tumors and sarcomas,\\
\rowcolor{Gray}\multirow{-2}{*}{\makecell{wechsler adult\\intelligence scale}}& nitrous\_oxide, psychiatric diagnosis, medical scales, level\_of\_measurement, adult, childhood\\

 & yin\_and\_yang, qi, alternative medicine, taoist cosmology, chinese martial arts terminology,\\
\multirow{-2}{*}{yang xiong} & chinese philosophy, plants used in traditional chinese medicine, gene, qigong, trees of china \\
\bottomrule
\end{tabular}
\caption{NED examples where dense BIER embeddings outperform sparse BIER representations. Mentions start with [N-Z].}
\label{tab:ned-de-not-sp-2}
\end{table*}


\begin{figure*}[t]
    \centering
    \includegraphics[width=1.0\linewidth]{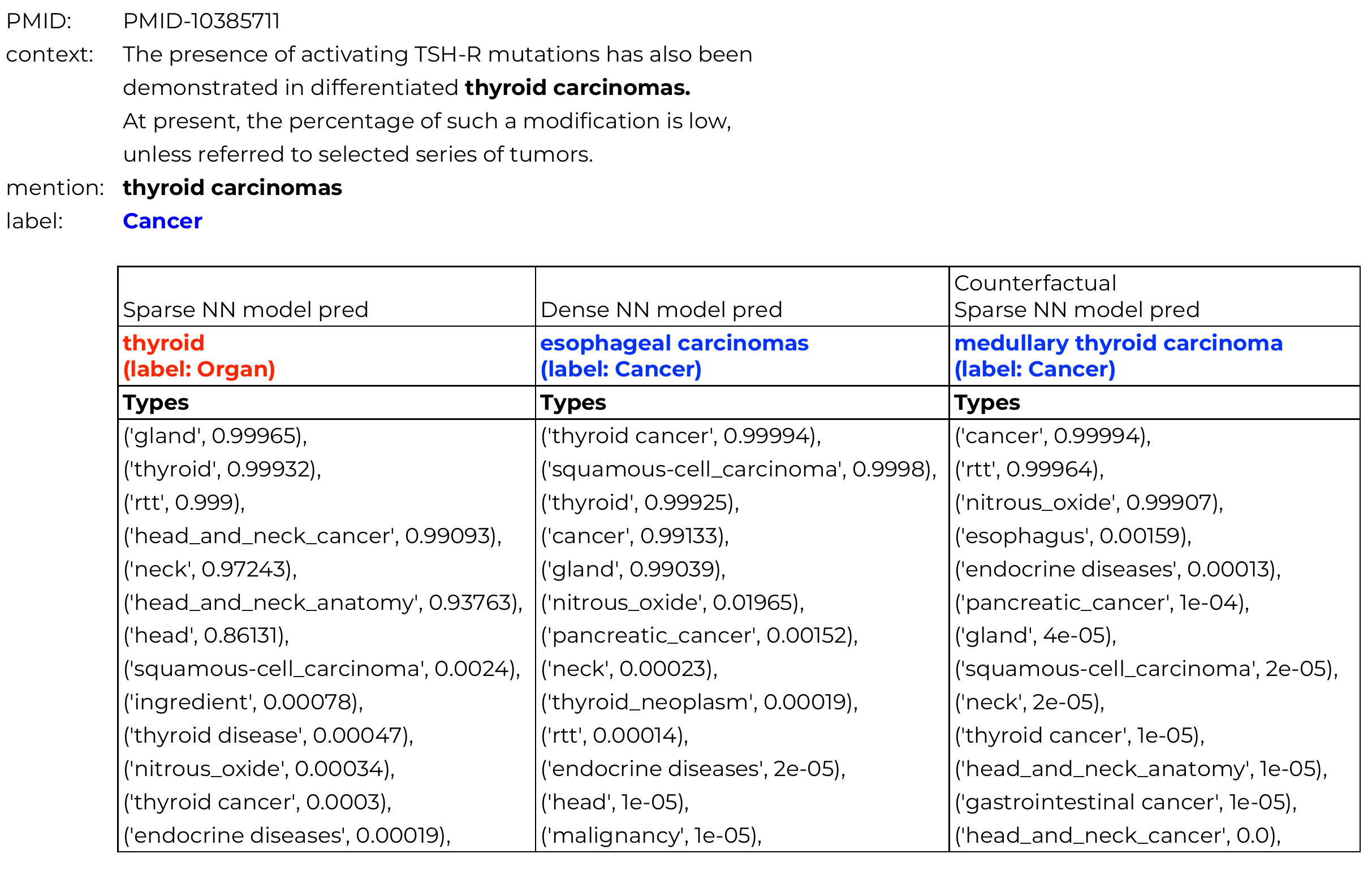}
    \caption{Analysis Example for ELC task}
    \label{fig:bier_analysis}
\end{figure*}


\end{document}